\pdfoutput=1
\documentclass[11pt]{article}

\usepackage[final]{acl}

\usepackage{times}
\usepackage{latexsym}
\usepackage{graphicx}
\usepackage{array}
\usepackage{booktabs}
\usepackage{graphicx, amsmath}
\usepackage{adjustbox}
\usepackage{caption}
\usepackage{multirow}
\usepackage{subcaption}
\usepackage{times}
\usepackage{latexsym}
\usepackage{algorithm}
\usepackage{algorithmicx}
\usepackage{algpseudocode}
\usepackage{amsmath}
\usepackage{amsfonts}
\usepackage{amssymb}
\usepackage{colortbl}
\usepackage{booktabs}
\usepackage{multirow}
\usepackage{adjustbox}
\usepackage{subcaption}
\usepackage{float}
\usepackage{algorithm}
\usepackage{algpseudocode} % or algorithmic, depending on what you're using
\usepackage[T1]{fontenc}
\usepackage[utf8]{inputenc}

\usepackage{microtype}

\usepackage{inconsolata}

\usepackage{graphicx}

\usepackage[most,skins,theorems]{tcolorbox}
\usepackage{pifont}

\tcbset{
  aibox/.style={
    width=\linewidth,
    top=8pt,
    bottom=2pt,
    colback=blue!6!white,
    colframe=black,
    colbacktitle=black,
    enhanced,
    center,
    attach boxed title to top left={yshift=-0.1in,xshift=0.15in},
    boxed title style={boxrule=0pt,colframe=white,},
  }
}
\newtcolorbox{AIbox}[2][]{aibox,title=#2,#1}

\newcommand{\bfit}[1]{\textbf{\textit{#1}}}

\title{Accelerating LLM Reasoning via Early Rejection with Partial Reward Modeling}

\author{
    \textbf{
    Seyyed Saeid Cheshmi\textsuperscript{1*}\hspace{3em}
    Azal Ahmad Khan\textsuperscript{1*}
    } \\
    \textbf{
    Xinran Wang\textsuperscript{1}\hspace{3em}
    Zirui Liu\textsuperscript{1}\hspace{3em}
    Ali Anwar\textsuperscript{1}
    } \\
    \textsuperscript{1}University of Minnesota\hspace{2em}\\
    \texttt{\{chesh014, khan1069, wang8740, zrliu, aanwar\}@umn.edu}
}

\begin{document}
\maketitle

\begin{abstract}
Large Language Models (LLMs) are increasingly relied upon for solving complex reasoning tasks in domains such as mathematics, logic, and multi-step question answering. 
A growing line of work seeks to improve reasoning quality by scaling inference time compute particularly through Process Reward Models (PRMs), used to reward the reasoning at intermediate steps. While effective, these methods introduce substantial computational overhead, especially when generating large numbers of solutions in parallel. 
In this paper, we investigate whether PRMs can be used mid-generation to provide early signals that enable the rejection of suboptimal candidates before full generation of step is complete. 
We introduce the hypothesis that PRMs are also Partial Reward Models, meaning that the scores they assign to partially completed reasoning step are predictive of final output quality. This allows for principled early rejection based on intermediate token-level signals. 
We support this hypothesis both theoretically, by proving that the risk of discarding optimal beams decreases exponentially with generation length and empirically, by demonstrating a strong correlation between partial and final rewards across multiple reward models. 
On math reasoning benchmarks, our method achieves up to 1.4$\times$–9$\times$ reduction in inference FLOPs without degrading final performance. 
These results suggest that early rejection is a powerful mechanism for improving the compute-efficiency of reasoning in LLMs. The code and implementation are available at \url{https://github.com/scheshmi/accelerated-reasoning-ER-PRM}.
\end{abstract}

\renewcommand{\thefootnote}{\fnsymbol{footnote}}
\footnotetext[0]{* Equal contributions (ordered via coin-flip).}
\renewcommand{\thefootnote}{\arabic{footnote}}

\section{Introduction}
\label{sec:Introduction}

% \ali{We bring in Process reward model without explaining what it is.}

% \ali{In abstract we bring in beam all of no where. Is beam well established? readers dont know what a beam is.}

% \azal{Updated the abstract to address both concerns.}

Large Language Models (LLMs) are at the forefront of AI capabilities due to their emerging ability to perform complex reasoning tasks~\cite{kojima2022large, chan2024understanding, cheng2025empowering, hazra2025have, xu2025towards}. They have demonstrated significant success in domains such as mathematical problem solving, multi-hop question answering, and logical inference~\cite{creswell2022selection, ahn2024large, akella2024improving}. These advancements are critical because they signal a shift from surface-level pattern recognition to deeper, multi-step deductive reasoning~\cite{wei2022chain, zhou2022least}. Enhancing these reasoning abilities is paramount for developing more capable, reliable models that can operate across various domains.

%%%%%%%%%%%%%%%%%%%%%%%%%%%%%%%%
\begin{figure*}
    \centering
    % \vspace{-2em}
    \includegraphics[width=0.8\linewidth]{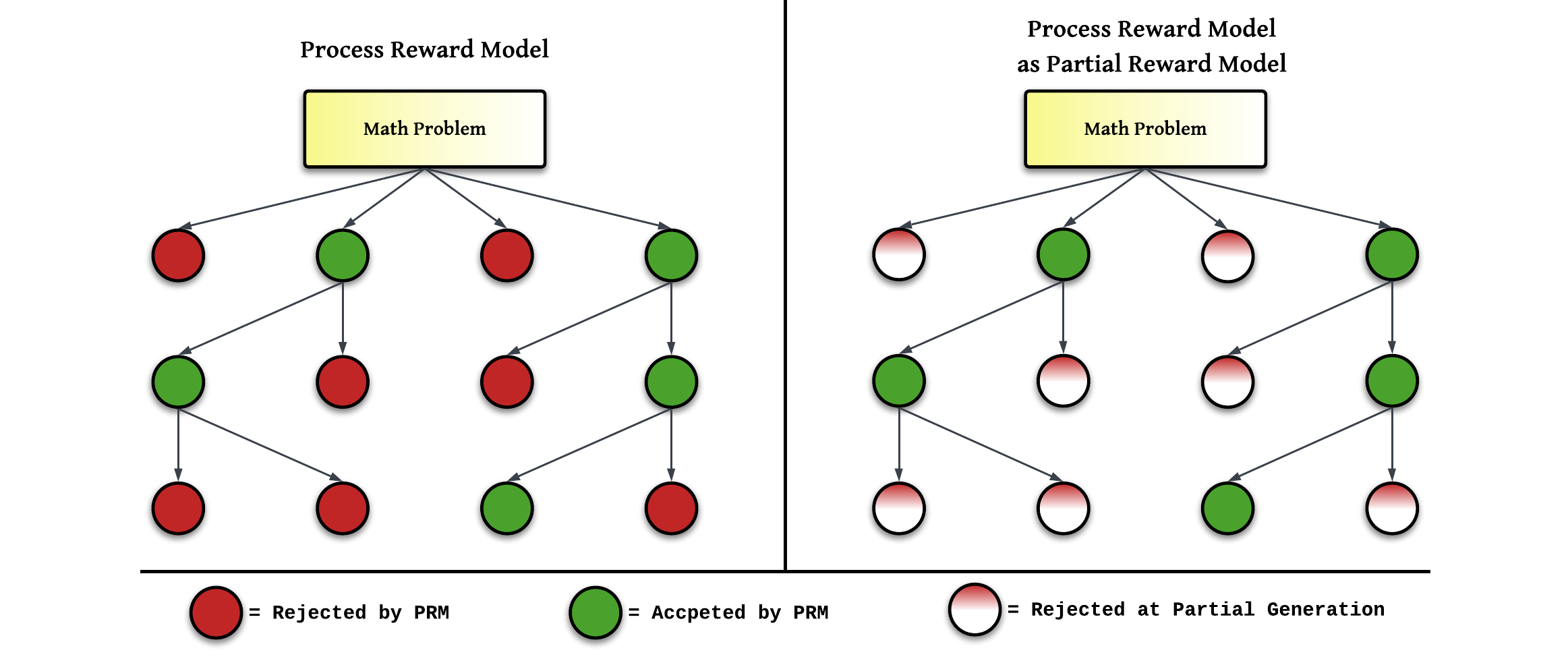}
    % \vspace{-2em}
    \caption{\textbf{Process‑Reward Model (PRM) at full length vs. PRM reused for early rejection.} \textbf{(Left)} Every beam is expanded to full depth before the PRM scores its complete solution, so all intermediate branches incur compute even if they are were to fail. \textbf{(Right)} The same PRM is invoked mid‑generation after each block of few tokens, producing a partial reward that serves as an early‑quality signal.}
    \label{fig:prm_diagram}
\end{figure*}
%%%%%%%%%%%%%%%%%%%%%%%%%%%%%%%%

%%%%%%%%%%% Plots %%%%%%%%%%%
  \begin{figure}[ht]
  \centering
  \begin{subfigure}[t]{0.225\textwidth}
    \centering
    \includegraphics[width=\linewidth]{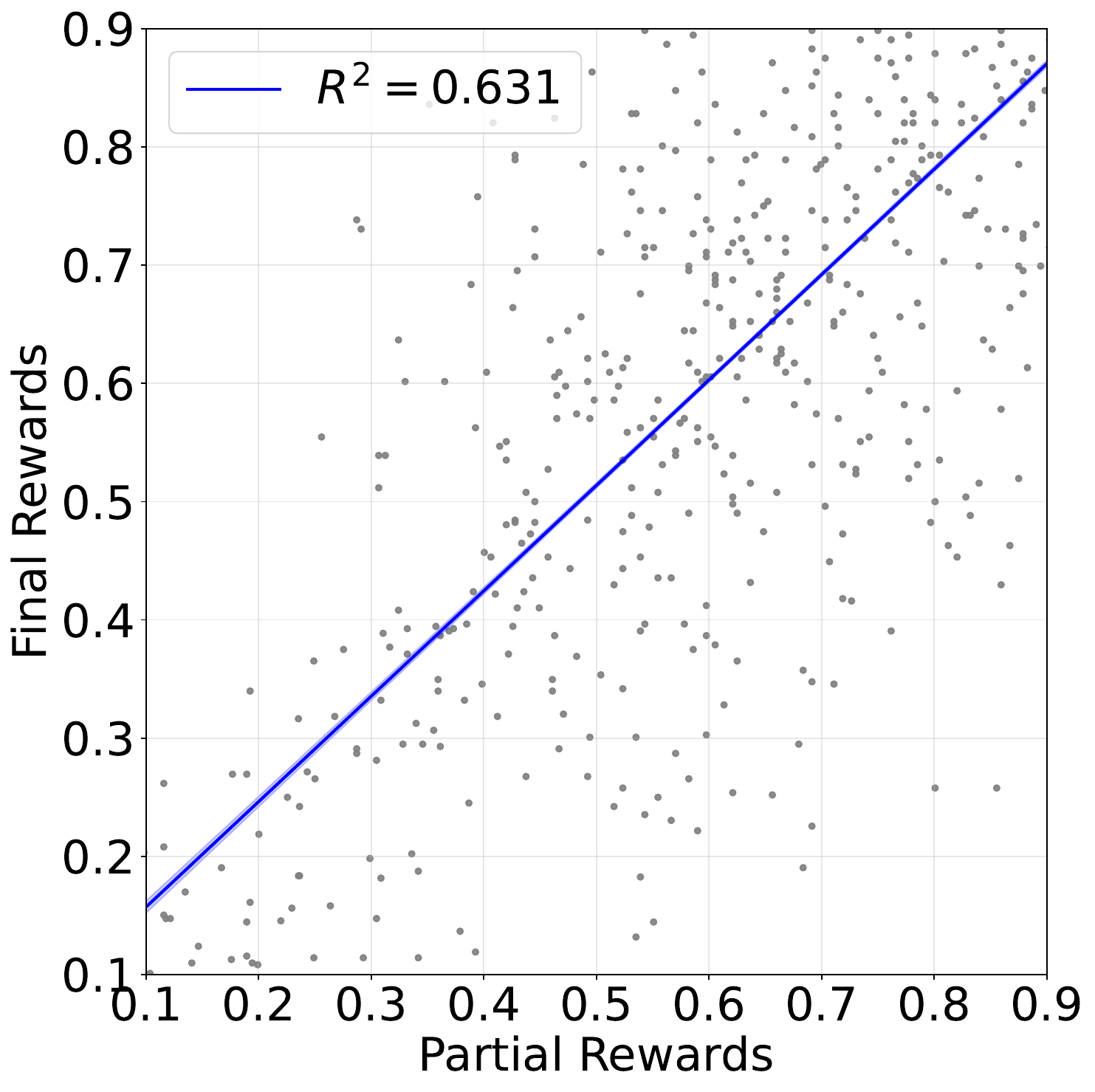}
  \end{subfigure}
  \quad % or another small horizontal space, if desired
  \begin{subfigure}[t]{0.225\textwidth}
    \centering
    \includegraphics[width=\linewidth]{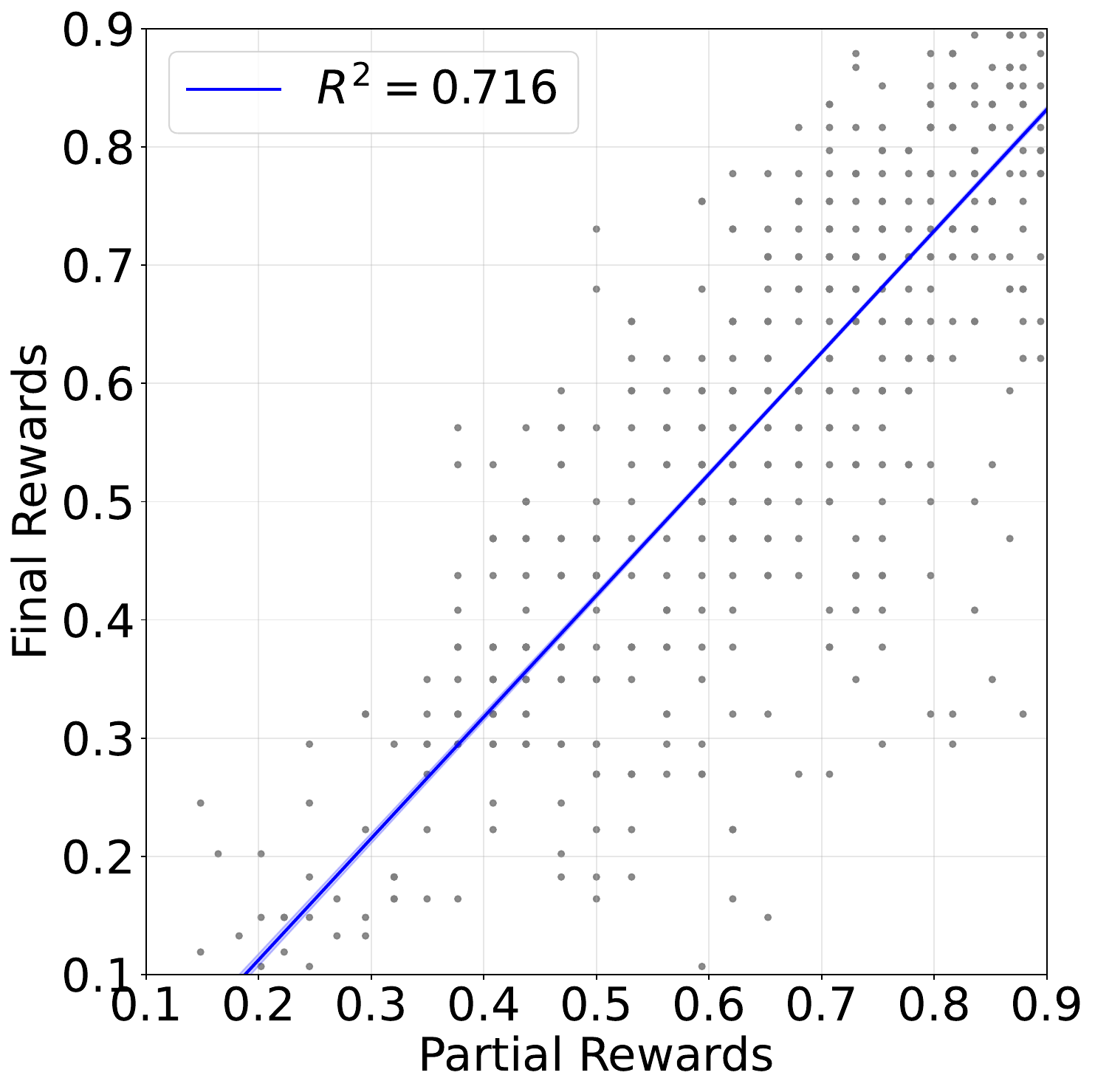}
  \end{subfigure}
  \caption{Linear relationship between partial rewards (reward calculated at half step completion) and full rewards (rewards calculated at full step completion) with \textbf{(top)} Llemma-MetaMath-7b and \textbf{(bottom)} MathShepherd-Mistral-7b as reward models.}
  \label{fig:linear_relationship}
  % \vspace{-1em}
\end{figure}
%%%%%%%%%%%%%%%%%%%%%%%%%%%%%%%%

\paragraph{Prior Works.}
As the scaling of model parameters and pretraining data has started to become a bottleneck, recent efforts have shifted toward increasing compute at inference time to improve the reasoning capabilities of LLMs~\cite{snell2024scaling}.
Improving the reasoning capabilities of LLMs by scaling compute at inference time has been pursued through multiple strategies. 
% A common method involves \emph{reasoning fine-tuned models}, where LLMs are trained on datasets explicitly curated to teach step-by-step reasoning processes~\cite{lobo2024impact, yu2025finemedlm}. 
A prominent approach leverages \emph{Outcome Reward Models}, which train a separate evaluator to score the final output of the LLM based on correctness or quality~\cite{cobbe2021training, hosseini2024v, mahan2024generative, zhang2024generative}. Another approach uses, \emph{Process Reward Models} (PRMs) to evaluate intermediate steps or reasoning trajectories generated during inference~\cite{wang2023math, snell2024scaling, zhang2024rest, luo2024improve}. 
Under this paradigm, the model generates multiple candidate reasoning paths, which are then evaluated by the PRM that assigns rewards at the end of each step. This step-wise evaluation enables the selection of promising trajectories for further expansion while allowing the rejection of less promising ones, thereby guiding the reasoning process more efficiently. Techniques such as beam search, Monte Carlo Tree Search (MCTS) guided by value models, and PRM-guided methods, exemplify this strategy~\cite{feng2023alphazero, yao2023tree}. In this paper, we focus specifically on the PRM paradigm and explore how to improve its efficiency.

\paragraph{Challenges.}
While scaling test-time compute using methods like PRMs can significantly enhance performance, it also introduces substantial computational overhead, especially for long sequences where many generated beams contribute little value~\cite{chen2024autoprm, hu2025prm, wang2025towards}. To be competitive with state-of-the-art post-training approaches, the number of beams often needs to be scaled up to 1000–60000, resulting in a large number of output tokens~\cite{sun2024fast}. These output tokens are not only computationally expensive to generate, but are also produced sequentially, leading to considerable latency~\cite{yang2024queueing}. A natural solution is to \emph{reject unpromising candidates early in the decoding process, after only a few tokens, before committing to full step generation}, a strategy we refer to as \emph{Early Rejection}.However, a major challenge in early rejection is making sure that decisions based on only part of the output don’t accidentally discard high-quality completions. This is difficult because the overall quality of a reasoning trace often depends on its full structure, which might not be obvious from the first few tokens. As a result, developing a reliable method that can make early yet accurate decisions about which traces to keep, based on partial generations, remains an important and open research problem.

\paragraph{Hypothesis.} To address this challenge, we present the a hypothesis, \bfit{Process Reward Models are also Partial Reward Models.} That is, for structured reasoning tasks, the partial scores assigned by a PRM, when evaluated after a small but meaningful fraction of the generation, are sufficiently correlated with the final scores. This insight suggests that PRMs, which are conventionally applied at the end of a complete reasoning trace, can also be used mid-generation to provide partial rewards that act as early indicators of output quality. Figure~\ref{fig:prm_diagram} illustrates this distinction: while traditional PRM usage scores complete reasoning paths only at the end, our approach invokes the same PRM mid-generation to score partial traces, enabling early rejection of unpromising candidates.
In doing so, they enable principled early rejection based on intermediate token-level signals. Preliminary results, as shown in Figure~\ref{fig:linear_relationship}, reveal a consistent relationship between partial and final rewards, modeled as a monotonic mapping with added noise.

\paragraph{Contributions.}
This paper makes the following key contributions:
\textbf{(C1)} \emph{We introduce the hypothesis that Process Reward Models (PRMs) can be used as Partial Reward Models to enable early rejection of suboptimal beams.} We support this hypothesis by showing that partial rewards computed after only a fraction of the generation are strongly correlated with final rewards, allowing for reliable early decisions in the reasoning process.
\textbf{(C2)} \emph{We provide theoretical guarantees that justify the use of partial scores for early rejection.} Specifically, we prove that under mild assumptions, the probability of prematurely rejecting the optimal trajectory decreases exponentially with the partial generation length.
\textbf{(C3)} \emph{We empirically demonstrate that early rejection guided by PRMs is both effective and compute-efficient.} On reasoning tasks such as AIME, Math-500 and AGI Eval, our approach reduces inference-time FLOPs by 1.4$\times$–9$\times$ when using a mid-sized PRM (7B parameters) without any loss in task performance. Furthermore, when using a smaller PRM (1.5B parameters), we achieve up to 1.5$\times$–4$\times$ reduction in FLOPs, demonstrating that even lightweight evaluators can enable highly efficient reasoning through early rejection.
\section{Related Works}
\label{sec:Related}

\paragraph{Generative Reward Models.}
Early approaches to guiding machine learning models relied on hand-crafted heuristics, but as models have grown more complex, generative models have increasingly been used for supervision and alignment~\cite{mahan2024generative}. Generative models now serve as critics, verifiers, and, most notably, as reward models in RLHF~\cite{mahan2024generative}. Critics evaluate model outputs by providing detailed feedback~\cite{luo2023critique, lan2024criticeval, lin2024criticbench, du2024llms}, while verifiers check the factual correctness or consistency of responses~\cite{kouemo2024chain, qi2024verifierq, kirchner2024prover}.
As reward models, they can be used to score either the final outcome (outcome reward models, ORMs) or provide feedback at intermediate steps (process reward models, PRMs)~\cite{lightman2024lets}. ORMs deliver a single reward signal at the end of generation, while PRMs offer denser, stepwise supervision, which has been shown to improve reasoning and generalization~\cite{cobbe2021training, wang2023math, hosseini2024v, zhang2024generative, snell2024scaling, luo2024improve}. 
PRMs have also been shown to facilitate more interpretable learning dynamics by providing actionable feedback at each reasoning step, enabling finer-grained control over model behavior and accelerating convergence during training~\cite{lightman2024lets, snell2024scaling, hosseini2024v}.
% Recent work demonstrates that generative reward models using self-generated reasoning traces and iterative optimization outperform classical methods, especially in out-of-distribution tasks~\cite{mahan2024generative}.

\paragraph{Early Rejection.}
In classification, confidence-based rejection and selective prediction methods~\cite{geifman19a} allow models to withhold outputs for ambiguous or out-of-distribution inputs, while similar abstention strategies are used in regression~\cite{mozannar20b}. In LLMs, early rejection began with Best-of-N (BoN) decoding, where all candidates are fully generated and only the best is selected~\cite{cobbe2021training, zhou2022least}. Recent advances show that integrating PRMs as step-level re-rankers within beam search significantly boosts both accuracy and compute efficiency, as dense rewards allow for rejection of suboptimal reasoning paths and more effective exploration of diverse solutions~\cite{wang2023math, snell2024scaling, zhang2024rest, luo2024improve}. Speculative Rejection proposed using ORMs for early rejection in BoN by discarding weak candidates mid-generation~\cite{sun2024fast}. In this work, \emph{we study the principle of early rejection for PRMs and demonstrate how it can be effectively integrated into beam search methods.}
\section{Method}
\label{sec:Method}

\paragraph{Beam Search for Reasoning.}
Beam search is a widely used decoding strategy in LLMs for structured generation tasks such as mathematical problem solving and multi-step reasoning~\cite{yao2023tree, feng2023alphazero, snell2024scaling}. At each decoding step, the model expands a fixed-width set of $N$ candidate beams by sampling multiple possible continuations and retaining only the top-scoring ones based on a predefined heuristic (e.g., log-probability or reward score). This iterative expansion and rejecting process allows the model to explore a larger space of possible outputs than greedy decoding, while remaining tractable compared to exhaustive search. In PRM-guided reasoning, each beam is scored at the end of every reasoning step by a PRM, which evaluates the coherence or correctness of the generated step. The highest scoring beams are then selected for further expansion, enabling the model to gradually construct a valid multi-step reasoning trace.

\subsection{Partial Scoring for Early Rejection}
\label{sec:subBackground}
Standard inference-time reasoning with LLMs and PRMs involves generating multiple candidate reasoning trajectories, typically using beam search or tree-based strategies, and scoring each trajectory after every step generation. Based on these step-wise scores, a subset of beams is selected and expanded further. While this strategy has been instrumental in advancing long-horizon reasoning, it incurs substantial computational overhead, as all candidate steps must be fully generated before evaluation, regardless of their quality.

We introduce a modification to this pipeline by reusing the same PRM mid-step generation. A compact overview is shown in Algorithm~\ref{fig:short-algorithm}, where instead of waiting for a full step to complete, we compute partial rewards after first block of $\tau$ tokens at each step. These intermediate scores serve as early indicators of downstream quality. Beams with low partial scores are rejected before completing the full step. The surviving beams are then completed to the end of the current step, after which expansion proceeds as in the standard pipeline. Early rejection is applied again at the next step. This process ensures that computation is focused on the most promising candidates, reducing the number of unnecessary tokens generated and minimizing redundant PRM evaluations. A full version of the algorithm, along with the standard PRM-guided baseline, is provided in Appendix~\ref{sec:Appendix} for reproducibility and implementation details.

% \begin{figure}[t]
%     \centering
%     \small
%     \begin{algorithm}[H]  % Or [tb], depending on your preference
%         \caption{Beam Search with Early Rejection}
%         \label{alg:partial}
%         \begin{algpseudocode}[1]
%             \Statex \textbf{Parameters:} 
%             \Statex \quad $N$: initial number of beams
%             \Statex \quad $M$: expansion factor
%             \Statex \quad $\tau$: partial rollout length
%             \Statex
%             \Function{BeamSearchEarlyRejection}{$N, M, \tau$}
%                 \State Initialize $N$ beams 
%                 \For{\texttt{each beam}}
%                     \State Generate up to $\tau$ tokens 
%                     \State Compute partial reward using PRM
%                 \EndFor
%                 \State Select top $\frac{N}{M}$ beams by partial reward
%                 \State Complete each selected beam to a full step
%                 \State Expand each selected beam into $M$ new beams
%                 \State Repeat partial scoring, early rejection, 
%                 \Statex \quad and expansion until stopping condition is met
%                 \State \Return \texttt{best final sequence}
%             \EndFunction
%         \end{algpseudocode}
%     \end{algorithm}
%     \caption{Overview of beam search with early rejection.}
%     \label{fig:short-algorithm}
% \end{figure}

\begin{figure}[htb]
    \centering
    \small
    \begin{algorithm}[H]
        \caption{Beam Search with Early Rejection}
        \label{alg:partial}
        \begin{algorithmic}[1]
            \State Initialize $N$ beams
            \For{each beam}
                \State Generate up to $\tau$ tokens and compute partial reward using PRM
            \EndFor
            \State Select top $N/M$ beams by partial reward and complete remaining beams to full step
            \State Expand each remaining beam with $M$ new beams
            \State Repeat scoring, early rejection, and expansion until stopping condition is met
            \State \Return Best final sequence
        \end{algorithmic}
    \end{algorithm}
    % \vspace{-2em}
    \caption{Overview of beam search with early rejection.}
    \label{fig:short-algorithm}
\end{figure}

\subsection{Efficiency Gains from Early Rejection}
\label{sec:Computational}

This early rejection strategy is focused on reduction in the number of tokens generated. By rejecting weaker candidates after a partial generation, we avoid expending compute on beams unlikely to contribute to the final output. The impact of this optimization on both generation cost and reward model evaluation is summarized below:

% \vspace{-0.5em}
\begin{AIbox}{Early rejection reduces compute}
Rejecting beams after generating first $\tau$ tokens leads to FLOPs reduction for each step generation and during PRM evaluation.
% Since most compute is concentrated in autoregressive decoding and PRM evaluation, early termination naturally leads to more efficient inference.
% Rejecting beams after $\tau$ tokens saves FLOPs by cutting short expensive decoding and reward evaluation, making inference significantly more efficient.
\end{AIbox}
% \vspace{-0.5em}

Beyond reducing total compute, early rejection also improves throughput through a two-tiered batching strategy. Since rejected beams only require $\tau$ tokens to be generated, they occupy significantly less memory. This enables to increase the batch size during the initial generation phase without getting OOM error. We then switch to a smaller batch size for completing the remaining beams, balancing exploration with memory efficiency. This batching decoupling is summarized below:

% \vspace{-0.5em}
\begin{AIbox}{Two-tiered batching improves throughput}
We use a larger batch size for generating the first $\tau$ tokens, taking advantage of their lower memory cost, and a smaller batch size for completing the step to avoid OOM error.
\end{AIbox}
% \vspace{-0.5em}
\section{Theoretical Guarantees}
\label{sec:Theory}

\paragraph{Background and Notation.}
At each decoding step, which we define as a block of $\tau$ tokens, a width of $N$ beams is maintained. For beam $i$, let $P_i$ denote its \emph{partial} reward after the first $\tau$ tokens and $F_i$ its \emph{final} reward after completing the step. Our preliminary results in Figure~\ref{fig:linear_relationship} indicate that the final reward is related to the partial reward via a monotonic mapping with added noise:

{
\small
\[
F_i = g(P_i) + \eta_i
\]
}
where $g$: $[0, 1] \xrightarrow{} [0, 1]$ is a monotonic increasing function; which need not be linear and $\eta_i$ is a noise term with zero mean and variance $\sigma^2$ that can cause deviations from a perfect linear relationship. After the PRM assigns partial rewards, we keep only the top $\frac{N}{M}$ beams and expand each of them into $M$ new beams, restoring the total width $N$. Let $p = \frac{N}{M}$, the selection threshold $T$ is the $(1-1/M)$ quantile of the partial-reward distribution (i.e., we keep the top $N/M$ beams). Therefore, a beam survives only if $P_i \ge T$.

Let the beam that would eventually yield the highest final score be

{
\small
\[
i^{*} \;=\; \arg\max_{i\in N} F_i
\]
}

% \paragraph{Guarantee Under Noisy Nonlinear Conditions.}
% Even though the relationship between $P_i$ and $F_i$ is not strictly linear, our assumption is that in expectation a higher partial score indicates a higher final score, and that the noise $\eta_i$ is "well-behaved" (for instance, sub-Gaussian with parameter $\sigma$). Let
% \[
% \Delta = \min_{j \ne i^*} (\mathbb{E[P_{i^*}]}-\mathbb{E[P_{j}]})
% \]
% denote the expected gap in the partial scores between the best beam $i^*$ and any other beam. We wish to bound the probability that the best beam is not in the top $p$ fraction. That is, we want to bound:
% \[
% Pr(P_{i^*} < T
% \]
% Under standard concentration arguments (e.g., using a Hoeffding or sub-Gaussian tail bound), if the noise $\eta_i$ is small relative to the expected gap $\Delta$, then the probability that $P_{i^*}$ falls below the $p$-th percentile is exponentially small. More precisely, one can derive a bound of the form:
% \[
% Pr(P_{i^*} < T) \le \exp (- \frac{(\Delta . (1-p))^2}{2 \sigma^2})
% \]
% This expression shows that if the gap $\Delta$ is large (i.e., the best beam is clearly ahead on average), and the retained fraction $p$ is not too small (so $1-p$ is significant), then the probability of mistakenly pruning the best beam is very low.

\paragraph{Guarantee under noisy, nonlinear conditions.}
Although the mapping between $P_i$ and $F_i$ need not be linear, we assume (i) the noise terms $\eta_i$ are independent and $\sigma$‑sub‑Gaussian, and (ii) the expected partial scores preserve the ordering of the expected final scores.  
Let

{\small
\[
\Delta \;=\; \min_{j\ne i^*} \bigl(\mathbb{E}[P_{i^*}] - \mathbb{E}[P_j]\bigr) \;>\; 0
\]
}
denote the smallest expected gap between the best beam $i^*$ and any other beam. 
% Because $T$ is the $(1/M)$‑quantile of the \emph{empirical} $\{P_j\}$, it is no larger than the same quantile of the expectations.
Thus
{\small
\[
\begin{split}
\Pr\bigl(P_{i^*}<T\bigr)
  &\le
  \Pr\Bigl(\exists\,j\ne i^*:\;P_j>P_{i^*}\Bigr) \\
  &\le
  (N-1)\,\exp\!\Bigl(-\tfrac{\Delta^{2}}{4\sigma^{2}}\Bigr),
\end{split}
\]
}
where the last step applies a sub‑Gaussian tail bound to each pairwise difference \(P_{i^*}-P_j\) and then takes a union bound over the \(N-1\) non‑optimal beams. The bound decays \emph{exponentially} in \(\Delta^{2}/\sigma^{2}\); thus, when the expected gap is appreciable and the noise is modest, the risk of pruning the optimal beam is negligible even for large beam widths.

%%%%%%%%%%%%%%%%%%%%%%%%%%%% Plots %%%%%%%%%%%%%%%%%%%%%%%%%%%%
% \begin{figure}[ht]
%   % \vspace{-2em}
%   \begin{subfigure}[t]{0.3\textwidth}
%     \centering
%     \includegraphics[width=\linewidth]
%     {Figures/kendall_correlation.pdf}
%   \end{subfigure}%
%   \hfill
%   \begin{subfigure}[t]{0.3\textwidth}
%     \centering
%     \includegraphics[width=\linewidth]{Figures/pearson_correlation.pdf}
%   \end{subfigure}%
%   \caption{\textbf{(Top)} Kendall's Tau and \textbf{(Bottom)} Pearson's correlation coefficient for the partial and final rewards.}
%   \label{fig:correlation}
%   \vspace{-1em}
% \end{figure}

\begin{figure}[ht]
  \centering
  \begin{subfigure}[t]{0.325\textwidth}
    \centering
    \includegraphics[width=\linewidth]{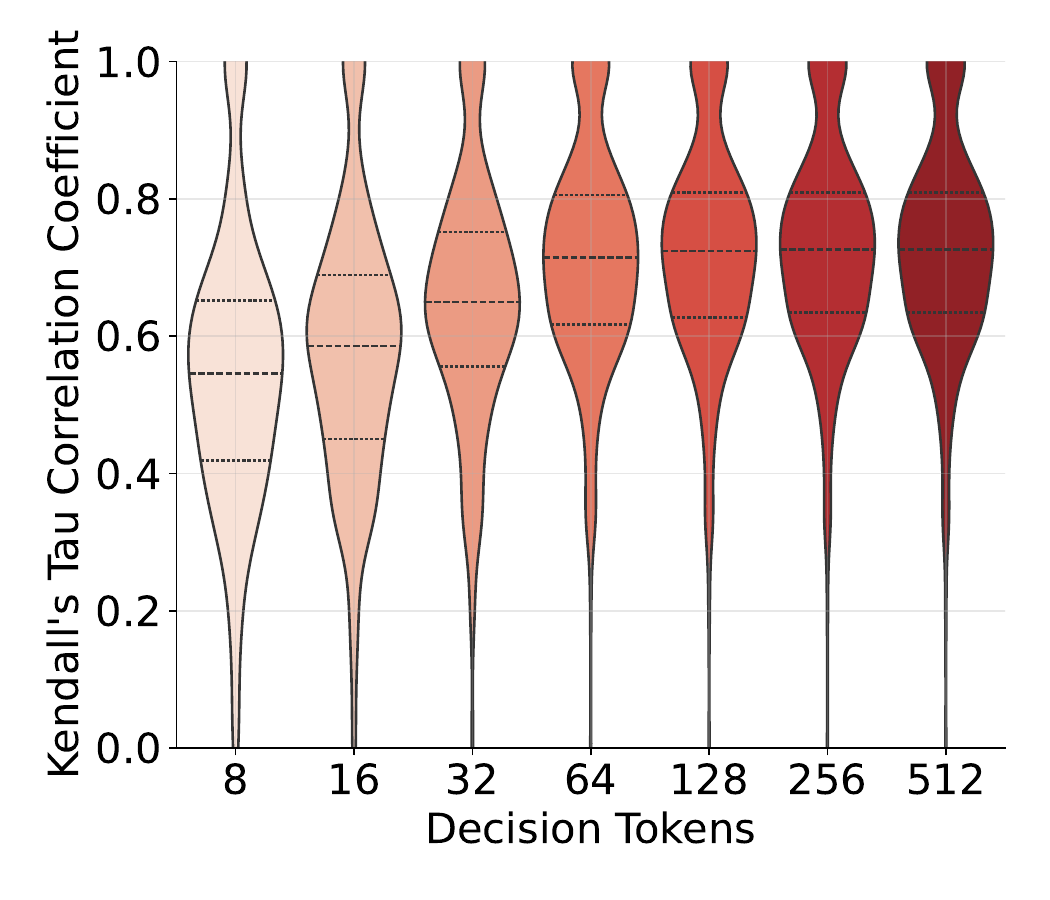}
  \end{subfigure}
  \quad % or another small horizontal space, if desired
  \begin{subfigure}[t]{0.325\textwidth}
    \centering
    \includegraphics[width=\linewidth]{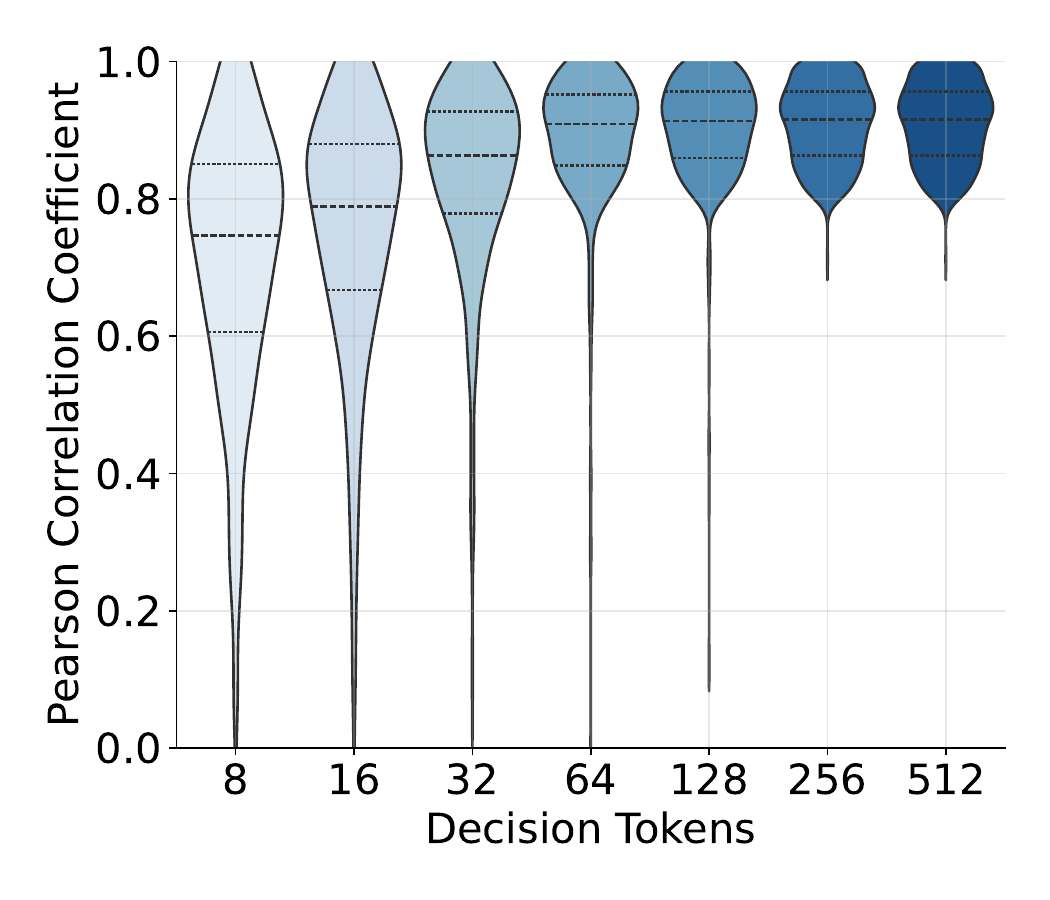}
  \end{subfigure}
  \caption{\textbf{(Top)} Kendall's Tau and \textbf{(Bottom)} Pearson's correlation coefficient for the partial and final rewards.}
  \label{fig:correlation}
  \vspace{-1em}
\end{figure}
%%%%%%%%%%%%%%%%%%%%%%%%%%%%%%%%%%%%%%%%%%%%%%%%%%%%%%%%%%%%%%%%%%%%

\paragraph{Best $\tau$ for Early Rejection.}
A common toy model is to treat each token’s (log‑)score as an i.i.d.\ random variable.  
For beam $i$, let $X_{i,1},\dots,X_{i,L}$ be i.i.d.\ with mean $\mu_i$ and variance $\sigma_i^{2}$,
where $L$ denotes the final sequence length (number of tokens at completion) and $1\le \tau \le L$.  
The partial reward after $\tau$ tokens is
$P_i=\sum_{t=1}^{\tau}X_{i,t}$, while the final reward is
$F_i=\sum_{t=1}^{L}X_{i,t}$.
Under this model the Pearson correlation reads

{\small
\[
\rho(P_i,F_i)=
\frac{\operatorname{Cov}(P_i,F_i)}
     {\sqrt{\operatorname{Var}(P_i)}\sqrt{\operatorname{Var}(F_i)}}
  \;=\;
  \sqrt{\frac{\tau}{L}}.
\]
}

The shared first $\tau$ tokens drive the entire covariance:  
as $\tau\!\to\!L$ the correlation approaches 1 meaning the partial score is an almost perfect proxy, whereas as $\tau\!\to\!0$ it vanishes.
Figure~\ref{fig:correlation} shows that this $\sqrt{\tau/L}$ trend, tightening toward 1 as $\tau$ increases, is also true empirically.

If we require the correlation to exceed a target level $\rho^{*}$, then

{\small
\[
\rho(P_i,F_i)=\sqrt{\tfrac{\tau}{L}}\;\ge\;\rho^{*}
\quad\Longrightarrow\quad
\tau\;\ge\;(\rho^{*})^{2}L.
\]
}

For example, attaining $\rho^{*}=0.8$ demands $\tau\ge0.64\,L$.

%%%%%%%%%%%%%%%%%%%%%%%%%%%%
\begin{figure*}
    \centering
    \includegraphics[width=0.75\linewidth]{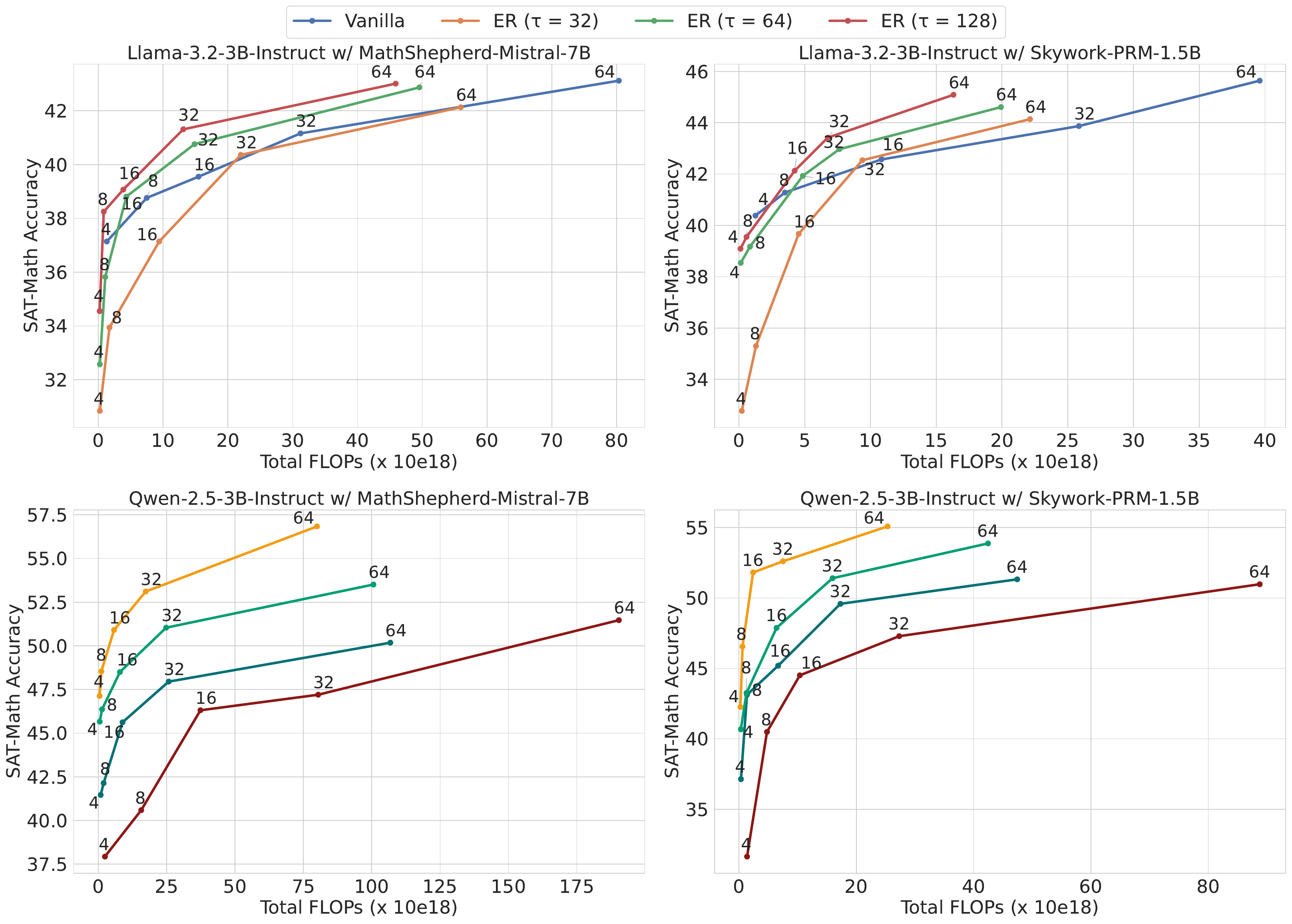}
    \caption{We evaluate our implementation of Early Rejection (ER) on the SAT-MATH dataset from AGIEval benchmark using two different LLMs and PRMs. The numbers indicate that ER rejection achieves performance similar to Vanilla Beam Search with while consuming far less compute.}
    \label{fig:satmath}
\end{figure*}
%%%%%%%%%%%%%%%%%%%%%%%%%%%%

\paragraph{Connection to the Sub‑Gaussian Bound.}
Our rejection guarantee hinges on

{\small
\[
\Pr\!\bigl(P_{i^*}<T\bigr)
   \;\le\; (N-1)\,
          \exp\!\Bigl(-\tfrac{\Delta^{2}}{4\sigma^{2}}\Bigr),
\]
}

where $\Delta=\min_{j\ne i^*}\bigl(\mathbb{E}[P_{i^*}]-\mathbb{E}[P_j]\bigr)$
is the expected partial‑score gap and $\sigma$ is the sub‑Gaussian
parameter of the per‑token noise.

A high correlation $\rho(P_i,F_i)$ does not automatically imply a large
gap $\Delta$, but it does indicate that beams ranking highly under the
partial reward tend to rank highly under the final reward.  In practice,
choosing $\tau$ so that

{\small
\[
\rho(P_i,F_i)=\sqrt{\tfrac{\tau}{L}}\;\ge\;\rho^{*}
\]
}
ensures the partial scores are sufficiently predictive; once this condition is met, the tail bound above tells us the probability of mistakenly pruning the optimal beam is exponentially small in \(\Delta^{2}/\sigma^{2}\). In practice, after fixing $\tau$ we measure the empirical gap $\Delta$ on a held-out set and confirm it comfortably exceeds the estimated noise scale~$\sigma$.
\section{Experiments}
\label{sec:Experiments}

% \begin{figure*}
%     \centering
%     \includegraphics[width=0.8\linewidth]{Figures/results.pdf}
%     \caption{We evaluate our implementation of Early Rejection (ER) on the SAT-MATH dataset from AGIEval benchmark using two different LLMs and PRMs. The numbers indicate that ER rejection achieves performance similar to Vanilla Beam Search with while consuming far less compute.}
%     \label{fig:satmath}
% \end{figure*}

We evaluate our method on three challenging math-reasoning benchmarks, MATH-500 \cite{lightman2023let}, SAT-MATH from AGIEval \cite{zhong2023agieval}, and AIME 2024.
For generation we use the instruct variants of two open-source LLMs, Llama-3.2-3B \cite{knuthwebsite} and Qwen-2.5-3B \cite{qwen2024qwen2}, selected for their strong reasoning ability at modest scale.
Process evaluation is performed with two PRMs of different capacities, MathShepherd-Mistral-7B \cite{wang2023math} and Skywork-PRM-1.5B, allowing us to study the impact of early rejection for PRMs of different sizes.
Early rejection is triggered after a prefix of $\tau \in {32, 64, 128}$ tokens. These thresholds are motivated by preliminary analysis (Figure \ref{fig:correlation}), which shows that partial-reward scores at these lengths are already highly correlated with final rewards. At each decoding step we sample $N \in {4, 8, 16, 32, 64}$ candidate beams and retain the top $M = 4$, mirroring prior PRM-guided search settings \cite{snell2024scaling}. We compare our early-rejection decoder with the conventional pipeline that scores only fully completed beams, reporting average answer accuracy and total inference FLOPs for each run.
All experiments are conducted on an HPC cluster, with each run executed using four NVIDIA A100 GPUs (40 GB memory each).

\subsection{Experimental Results}

\begin{figure*}
    \centering
    \includegraphics[width=0.75\linewidth]{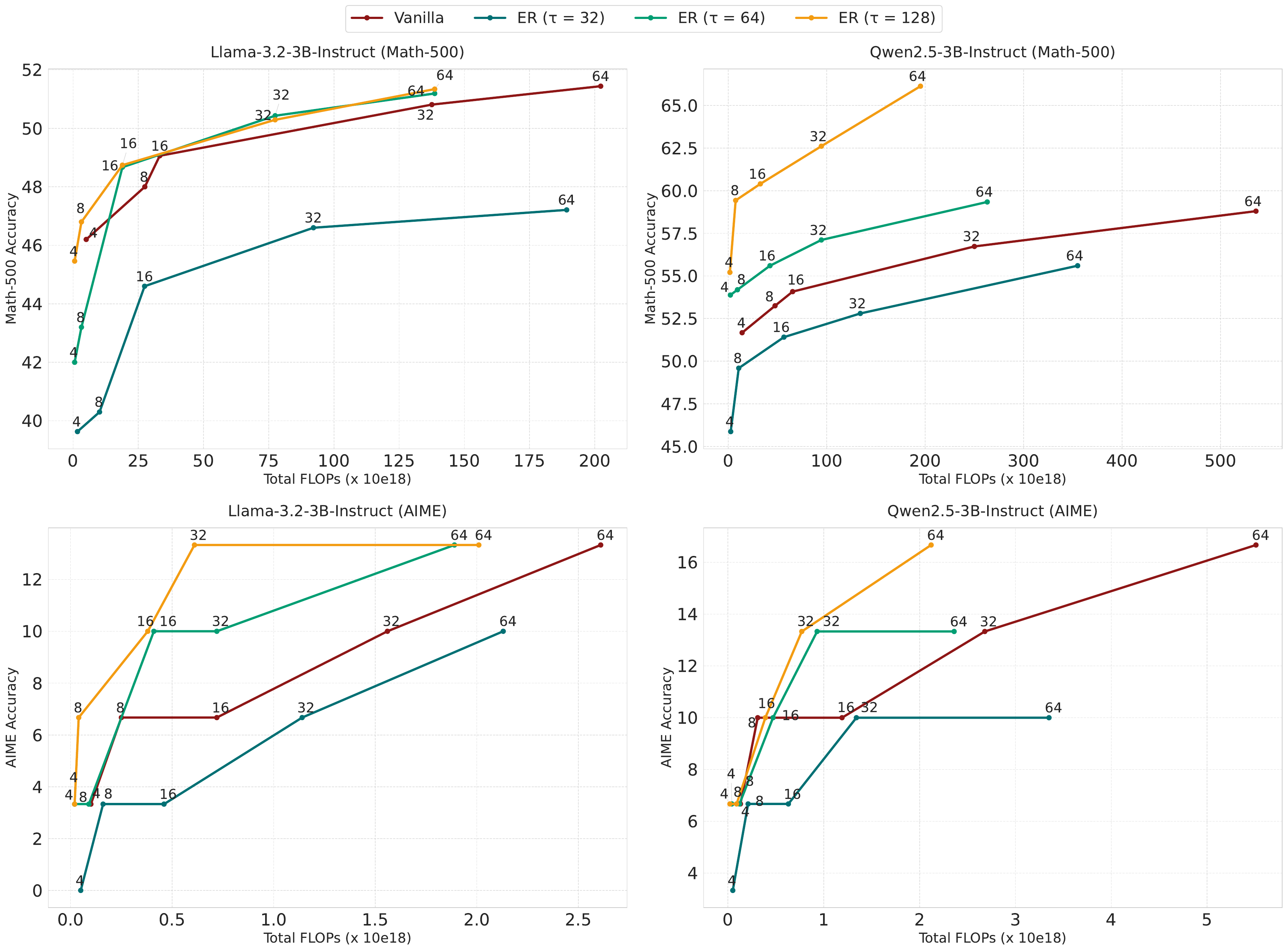}
    \caption{We evaluate our implementation of Early Rejection (ER) on the Math-500 and AIME datasets using two different LLMs with MathShepard-7b as reward model. The numbers indicate that ER rejection achieves performance similar to Vanilla Beam Search with while consuming far less compute.}
    \label{fig:math500}
\end{figure*}

Experimental results in Figure~\ref{fig:satmath} on SAT-MATH dataset and~\ref{fig:math500} on Math-500 and AIME 2024 datasets highlight the effectiveness of Early Rejection (ER) in reducing compute while preserving end-task accuracy across different PRMs, LLMs, and $\tau$ values. For the results we observe that across all configurations, early rejection acts as a safe and compute-efficient strategy that adapts well to LLM characteristics and PRM granularity. 
Appendix~\ref{sec:Appendix} provides a comprehensive breakdown of accuracy and compute trade-offs across all datasets, $\tau$ values, beam sizes, and LLM–PRM configurations.
Building on these results, we articulate five key observations that our subsequent experiments directly address.

% \noindent \textbf{RQ1: How strongly do partial PRM scores at different token lengths correlate with final PRM scores?}

\textbf{Observation \ding{182}: Partial PRM scores at very short prefixes reliably predict final scores.}
Our empirical analysis confirms that partial rewards become highly predictive of final rewards after surprisingly short prefixes. Figure~\ref{fig:correlation} shows as we sweep the decision prefix $\tau$ from 8 to 512 tokens. The two correlations rise monotonically and follow the $\sqrt{\tau/L}$ and at $\tau$ = 32 tokens $\rho$ already exceeds 0.78 and $\tau$ = 64 pushes both metrics above 0.9, after which they plateau. A complementary view is given in Figure~\ref{fig:linear_relationship}, where a linear fit between half-step partial rewards and full-step rewards achieves $R^2$ = 0.63 with the MetaMath-7B PRM and $R^2$ = 0.72 with MathShepherd-7B, demonstrating that the effect generalizes across reward models . These findings validate our Partial Reward Model hypothesis that even a one-third length prefix offers a stable ranking signal, and the probability of incorrectly rejecting the optimal beam decays exponentially once the expected partial-score gap $\Delta$ dominates the sub-Gaussian noise $\sigma$, as formalized in Section~\ref{sec:Theory}. Practically, this means we can invoke early rejection after the first 32–64 tokens with negligible risk while removing 60–85\% of downstream PRM calls and generation FLOPs.

% \vspace{-0.5em}
% \begin{AIbox}{Partial Scoring Saves Compute}
% Using partial rewards for early rejection reduces 60–85\% of FLOPs while retaining accuracy within 0.3–0.7\% of full-step inference, validating its practical utility.
% \end{AIbox}
% \vspace{-0.5em}

\textbf{Observation \ding{183}: Smaller PRMs can match or exceed larger PRMs in accuracy while saving more compute, especially on well‐structured outputs.}
The smaller Skywork-PRM-1.5B achieves equal or higher end-task accuracy than the MathShepherd-Mistral-7B baseline, while also enabling a higher number of FLOP reductions. Across both Llama-3.3-3B and Qwen2.5-3B, Skywork yields a 0.7–2.1\% accuracy gain for smaller beam sizes and stays within 0.3\% elsewhere, contradicting the common intuition that \emph{``bigger judge = better answers”}~\cite{leike2018scalable}. We also observe a greater number of FLOP reductions with Skywork-PRM-1.5B, primarily because the 3B-sized LLM becomes the computational bottleneck, and early rejection allows us to skip costly completions, thereby saving compute more frequently.

Another key observation is that smaller PRMs benefit from more well-structured answers. Skywork-PRM-1.5B generally performs better with Llama-3.3-3B than with Qwen2.5-3B, as Llama tends to produce more structured and instruction-following responses compared to Qwen. Although both LLMs are instruction-tuned, Llama adheres to instructions more faithfully, making it easier for the smaller PRM (Skywork) to evaluate intermediate steps accurately. In contrast, larger PRMs like MathShepherd-Mistral-7B are more robust to such variations in LLM behavior.
 
% \vspace{-0.5em}
% \begin{AIbox}{Smaller PRMs Prefer Structured Outputs}
% Smaller PRMs assign more reliable partial rewards when intermediate reasoning traces are presented in well-structured format, enabling more accurate early rejection.
% \end{AIbox}
% \vspace{-0.5em}

\begin{figure*}
    \centering
    \includegraphics[width=\linewidth]{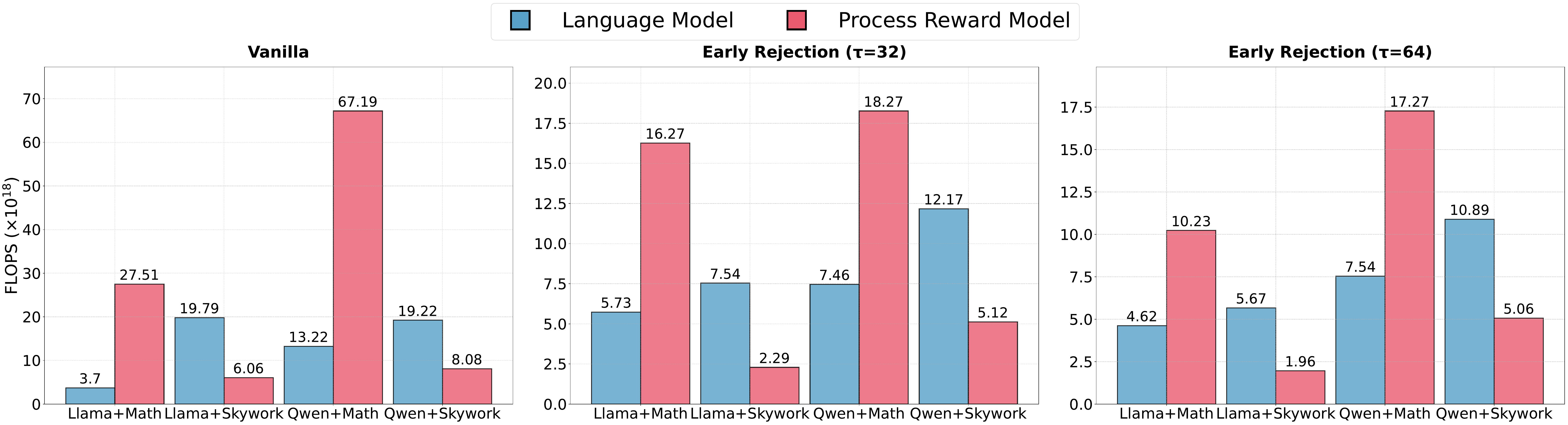}
    \caption{Total FLOPs consumed across different LLM–PRM combinations with and without Early Rejection. We observe consistent and substantial reductions in compute, with $\tau=$ 64 yielding up to 9× savings. Larger prefix lengths enable more reliable pruning, significantly lowering overall inference cost without compromising accuracy.}
    \label{fig:flopeval}
\end{figure*}

\textbf{Observation \ding{184}: Early rejection yields large accuracy gains for exploratory LLMs at small beam widths but offers diminishing accuracy returns for deterministic LLMs and wider beams.}
Qwen-2.5B often generates long, exploratory reasoning traces, so many beams appear weak after the first $\tau = 32$–$64$ tokens, even though some of them would eventually reach correct solutions. In such cases, the partial reward filter discards the clearly unpromising beams early. Here early rejection frees up beam slots for new candidates. This allows the search to explore a broader set of reasoning paths, effectively expanding the search space without increasing the beam width $N$.

In contrast, Llama-3.2-3B tends to produce shorter, more deterministic traces where the top-$p$ beams already rank highly from the start. As a result, early rejection removes fewer low-quality candidates and provides limited additional exploration. Empirically, early rejection improves Qwen's accuracy by up to 3.5\% at $N = 4$ and 1.6\% at $N = 8$, whereas Llama sees at most a 0.3\% gain. Once the beam width is sufficiently large ($N \geq 32$), the baseline search already explores the space well, so the benefits of early rejection shift from accuracy gains to compute savings.

\textbf{Observation \ding{185}: At $\tau = 64$ tokens, early rejection achieves higher accuracy while lesser compute than $\tau = 32$ tokens.}  
Although we always retain the same number of beams per step (the top $N/M$), their quality improves significantly as we increase the prefix length $\tau$. At $\tau = 32$, the correlation between partial and full rewards is about 0.78. This means around 20\% of the beams are incorrectly ranked, so some low-quality beams make it through and have to be fully generated and evaluated, wasting compute.
At $\tau = 64$, the correlation exceeds 0.90 and flattens out, meaning nearly all retained beams are genuinely promising. Very few low-quality beams slip through. As a result, even though we keep the same number of survivors, the number of bad survivors and the FLOPs spent on them, drops when increasing $\tau$ from 32 to 64.

\textbf{Observation \ding{186}: Language model behavior (not size) drives compute, and early rejection is most effective when it blocks exploratory failures early.}
Figure~\ref{fig:flopeval} shows that Qwen2.5-3B incurs significantly higher total FLOPs than Llama-3.2-3B under identical early rejection settings. While both models are similar in size, their generation behaviors differ: Qwen tends to produce longer, exploratory chains of thought, whereas Llama generates more concise, deterministic outputs. As a result, when early rejection fails to prune a weak Qwen beam, it often leads to a long and costly completion, inflating total compute.
Early rejection is most effective in these exploratory settings, where catching bad completions early prevents large downstream FLOPs. This explains why Qwen exhibits larger absolute FLOP reductions, especially when paired with a lightweight PRM like Skywork-1.5B. In contrast, Llama's beams tend to converge quickly, offering fewer opportunities for savings. These results highlight that the structure of the generation process, not just model size, governs the impact of early rejection on efficiency.

% \vspace{-0.5em}
% \begin{AIbox}{Early Rejection helps Exploratory Models}
% Qwen benefits from early rejection at small $N$ due to its exploratory reasoning, while Llama's deterministic paths already rank well and gain little from resampling.
% \end{AIbox}
% \vspace{-0.5em}

% \input{Figures/table_1}
% \input{Figures/table_2}

\section{Conclusion}
\label{sec:Conclusion}
We demonstrate that PRMs can be effectively repurposed as Partial Reward Models, enabling a single mid-generation evaluation to provide a reliable accept or reject signal. This allows weak beams to be pruned early, well before full reasoning steps are completed, thereby reducing unnecessary computation without sacrificing final accuracy.
Under mild noise assumptions, we provide theoretical guarantees showing that the probability of mistakenly discarding the optimal beam decays exponentially with prefix length, offering formal safety for early rejection.
Extensive experiments across SAT-MATH, Math-500, and AIME confirm the practical benefits: early rejection reduces inference-time FLOPs by 1.4$\times$–9$\times$ when using a mid-sized PRM (7B parameters), with no degradation in task performance. Even with a smaller PRM (1.5B), we observe 1.5$\times$–4$\times$ compute savings, highlighting that lightweight evaluators are sufficient for effective and efficient reasoning.
Together, these findings establish early rejection as a simple, model-agnostic plug-in that narrows the gap between compute-heavy tree search and fast single-pass decoding, offering state-of-the-art compute efficiency without compromising solution quality.
\section*{Limitations}
\label{sec:Limitations}

Our approach relies on the monotonicity and calibration of PRM scores, if partial rewards correlate weakly with final quality, as might occur in tasks with delayed or non-monotonic utilities (e.g., code synthesis with backtracking or creative writing), early rejection can mis-reject the eventual best beam. The study is confined to text-only, math-centric benchmarks. Larger models specially for multimodal tasks, or domains with sparse positive signals may exhibit different trade-offs. While we report FLOP reductions, we do not quantify the memory overhead of storing intermediate PRM states after $\tau$ tokens are generated. Finally, the theoretical guarantees assume independent step-wise noise and fixed $\tau$, leaving open questions about adaptive $\tau$ schedules and integration with policy-learning frameworks such as RLHF or DPO.
\section*{Ethical Considerations}
\label{sec:Ethical_Considerations}
% Early‑Rejection shrinks inference compute by up to 9$\times$, sharply lowering the hardware barrier for mass‑producing multi‑step outputs, a benefit that could equally enable large‑scale spam or disinformation. Its safety guarantees hinge on PRM scores remaining monotonic with final quality, but the authors concede this may fail outside the paper's math‑centric tests (e.g., tasks with delayed or non‑monotonic rewards), so the algorithm might prematurely drop the best answer or encode unseen biases.

While Early Rejection reduces inference compute by up to 9$\times$, this efficiency could also facilitate misuse, such as the automated generation of spam or disinformation. The method's safety relies on the assumption that PRM scores are monotonic with respect to final output quality. However, this assumption may not hold beyond the math-focused benchmarks evaluated in this work, particularly for tasks involving delayed or non-monotonic rewards. As a result, the algorithm risks discarding high-quality candidates prematurely or reinforcing hidden biases.

\section*{Acknowledgments}
The work of Azal Ahmad Khan was supported in part by the Amazon Machine Learning Systems Fellowship and the UMN GAGE Fellowship.
Xinran Wang and Ali Anwar were supported by the 3M Science and Technology Graduate Fellowship and the Samsung Global Research Outreach Award.

\bibliography{bibliography}

% \newpage
\clearpage
\onecolumn

\appendix
\section{Appendix}
\label{sec:Appendix}

\begin{figure}[htb]
    \centering
    \small % adjust font size
    \begin{algorithm}[H]
        \caption{Beam Search with Process Reward Models}
        \label{fig:beam_search_prm}
        \begin{algorithmic}[1]
            \State \textbf{Input:} LLM Model, PRM Model, Beam count $N$, Beam width $M$, Temperature $T$, Stopping criterion, EOS token or max search depth, Batch Size $b$
            \State Initialize a set of $N$ candidate beams
            \For{each beam}
                \State Sample $N$ independent steps using the LLM with temperature $T$
                \State Apply the stopping criterion (e.g., new line or double new line)
            \EndFor
            \State Score each sampled step using the PRM
            \State Select the top $N/M$ steps with the highest scores
            \State Expand the selected steps:
            \For{each selected step}
                \State Sample $M$ next steps
            \EndFor
            \While{EOS token not reached \textbf{and} max search depth not exceeded}
                \State Repeat steps 7 - 12
            \EndWhile
            \State \Return The best sequence found
        \end{algorithmic}
    \end{algorithm}
\end{figure}

\begin{figure}[htb]
    \centering
    \small % adjust font size
    \begin{algorithm}[H]
        \caption{Beam Search with Early Rejection}
        \label{fig:beam_search_rejection}
        \begin{algorithmic}[1]
            \State \textbf{Input:} LLM Model, PRM Model, Beam count $N$, Beam width $M$, Temperature $T$, Stopping criterion, EOS token or max search depth, $b_1 > b_2$
            \State Initialize a set of $N$ candidate  beams
            \For{each beam}
                \State Sample $N$ independent steps using the LLM with temperature $T$ and batch size $b_1$
                \State Apply the stopping criterion ($\tau$ tokens generated or EOS token.)
            \EndFor
            \State Score each sampled step using the PRM
            \State Select the top $N/M$ steps with the highest scores
            \State Complete the selected steps:
            \For{each selected step}
                \State Complete the step until EOS token with batch size $b_2$.
            \EndFor
            \State Expand the selected steps:
            \For{each selected step}
                \State Sample $M$ next steps
            \EndFor
            \While{EOS token not reached \textbf{and} max search depth not exceeded}
                \State Repeat steps 7 - 16
            \EndWhile
            \State \Return The best sequence found
        \end{algorithmic}
    \end{algorithm}
\end{figure}

\paragraph{Algorithm.}
Algorithm~\ref{fig:beam_search_prm} shows the conventional PRM-guided beam search and Algorithm~\ref{fig:beam_search_rejection} shows our early-rejection variant. Both algorithms maintain the same top-level structure of iterative beam expansion, but differ critically in how and when PRM scores are computed. The standard method evaluates only fully completed beams, resulting in redundant computation on unpromising candidates. In contrast, our early-rejection variant computes partial rewards after just $\tau$ tokens using the same PRM, enabling efficient early rejecting. This architectural shift introduces a two-tiered batching scheme, larger batch size for partial generations and smaller batch size for step completion, yielding significant compute savings without degrading performance, as shown in our experimental results.

\paragraph{Results.}
To supplement the main results presented in Section 5, we provide detailed tables reporting the accuracy and compute trade-offs for every combination of language model (LLM), process reward model (PRM), beam size, and early rejection threshold $\tau$. These results span three math reasoning benchmarks: SAT-MATH, Math-500, and AIME.

Table~\ref{tab:SAT_Math_performance} reports the results on the SAT-MATH dataset from AGIEval. For each LLM–PRM pair, we compare standard decoding ("Vanilla") with our early rejection method across multiple $\tau$ values. Each cell reports both the accuracy and the total FLOPs used for inference. We observe that early rejection achieves similar or higher accuracy at significantly reduced compute, especially with exploratory LLMs like Qwen-2.5B.

\begin{table}[htbp]
\centering
\caption{SAT-MATH results comparing vanilla decoding and Early Rejection (ER) across multiple beam sizes and $\tau$ values. Each cell reports (top) accuracy and (bottom) total inference FLOPs ($\times10^{18}$).}
\label{tab:SAT_Math_performance}
\begin{adjustbox}{max width=\textwidth}
\begin{tabular}{@{}l|l|l|*{5}{c}@{}}
\toprule
\multirow{2}{*}{\textbf{Model}} & \multirow{2}{*}{\textbf{PRM}} & \multirow{2}{*}{\textbf{Setting}} & \multicolumn{5}{c}{\textbf{Number of Samples ($\tau$)}} \\
\cmidrule(lr){4-8}
& & & \textbf{4} & \textbf{8} & \textbf{16} & \textbf{32} & \textbf{64} \\
\midrule

% Llama-3.2-3b with MathSheperd-7b
\multirow{8}{*}{\begin{tabular}[c]{@{}c@{}}\textbf{Llama-3.2}\\\textbf{-3b}\end{tabular}} 
& \multirow{8}{*}{\begin{tabular}[c]{@{}c@{}}\textbf{MathSheperd}\\\textbf{-7b}\end{tabular}} 
& \textit{Vanilla} & 37.14 & 38.76 & 39.55 & 41.16 & 43.12 \\
& & & \textcolor{gray}{\footnotesize 1.32} & \textcolor{gray}{\footnotesize 7.48} & \textcolor{gray}{\footnotesize 15.47} & \textcolor{gray}{\footnotesize 31.21} & \textcolor{gray}{\footnotesize 80.34} \\
\cmidrule(lr){3-8}
& & \textit{ER ($\tau$ = 32)} & 30.84 & 33.94 & 35.14 & 40.36 & 42.13 \\
& & & \textcolor{gray}{\footnotesize 0.24} & \textcolor{gray}{\footnotesize 2.73} & \textcolor{gray}{\footnotesize 9.40} & \textcolor{gray}{\footnotesize 21.99} & \textcolor{gray}{\footnotesize 55.94} \\
\cmidrule(lr){3-8}
& & \textit{ER ($\tau$ = 64)} & 32.57 & 35.82 & 38.81 & 40.76 & 42.87 \\
& & & \textcolor{gray}{\footnotesize 0.24} & \textcolor{gray}{\footnotesize 1.08} & \textcolor{gray}{\footnotesize 4.34} & \textcolor{gray}{\footnotesize 14.85} & \textcolor{gray}{\footnotesize 49.55} \\
\cmidrule(lr){3-8}
& & \textit{ER ($\tau$ = 128)} & 34.55 & 38.25 & 39.07 & 38.31 & 40.65 \\
& & & \textcolor{gray}{\footnotesize 0.21} & \textcolor{gray}{\footnotesize 0.85} & \textcolor{gray}{\footnotesize 3.86} & \textcolor{gray}{\footnotesize 13.11} & \textcolor{gray}{\footnotesize 45.90} \\

\cmidrule(lr){2-8}

% Llama-3.2-3b with Skywork-1.5b
& \multirow{8}{*}{\begin{tabular}[c]{@{}c@{}}\textbf{Skywork}\\\textbf{-1.5b}\end{tabular}} 
& \textit{Vanilla} & 40.38 & 41.28 & 42.57 & 43.87 & 45.64 \\
& & & \textcolor{gray}{\footnotesize 1.25} & \textcolor{gray}{\footnotesize 3.49} & \textcolor{gray}{\footnotesize 10.83} & \textcolor{gray}{\footnotesize 25.85} & \textcolor{gray}{\footnotesize 39.60} \\
\cmidrule(lr){3-8}
& & \textit{ER ($\tau$ = 32)} & 32.77 & 35.30 & 39.67 & 38.54 & 44.14 \\
& & & \textcolor{gray}{\footnotesize 0.21} & \textcolor{gray}{\footnotesize 1.29} & \textcolor{gray}{\footnotesize 4.54} & \textcolor{gray}{\footnotesize 9.38} & \textcolor{gray}{\footnotesize 22.13} \\
\cmidrule(lr){3-8}
& & \textit{ER ($\tau$ = 64)} & 38.54 & 39.17 & 41.93 & 42.97 & 44.61 \\
& & & \textcolor{gray}{\footnotesize 0.13} & \textcolor{gray}{\footnotesize 0.83} & \textcolor{gray}{\footnotesize 4.85} & \textcolor{gray}{\footnotesize 7.63} & \textcolor{gray}{\footnotesize 19.92} \\
\cmidrule(lr){3-8}
& & \textit{ER ($\tau$ = 128)} & 32.24 & 33.33 & 37.21 & 39.09 & 39.55 \\
& & & \textcolor{gray}{\footnotesize 0.11} & \textcolor{gray}{\footnotesize 0.57} & \textcolor{gray}{\footnotesize 4.23} & \textcolor{gray}{\footnotesize 6.75} & \textcolor{gray}{\footnotesize 16.31} \\

\midrule

% Qwen2.5-3b with MathSheperd-7b
\multirow{8}{*}{\begin{tabular}[c]{@{}c@{}}\textbf{Qwen2.5}\\\textbf{-3b}\end{tabular}} 
& \multirow{8}{*}{\begin{tabular}[c]{@{}c@{}}\textbf{MathSheperd}\\\textbf{-7b}\end{tabular}} 
& \textit{Vanilla} & 37.93 & 40.59 & 46.31 & 47.20 & 51.47 \\
& & & \textcolor{gray}{\footnotesize 2.42} & \textcolor{gray}{\footnotesize 15.70} & \textcolor{gray}{\footnotesize 37.35} & \textcolor{gray}{\footnotesize 80.41} & \textcolor{gray}{\footnotesize 190.35} \\
\cmidrule(lr){3-8}
& & \textit{ER ($\tau$ = 32)} & 41.46 & 42.14 & 45.62 & 47.95 & 50.18 \\
& & & \textcolor{gray}{\footnotesize 0.86} & \textcolor{gray}{\footnotesize 1.96} & \textcolor{gray}{\footnotesize 8.85} & \textcolor{gray}{\footnotesize 25.73} & \textcolor{gray}{\footnotesize 106.77} \\
\cmidrule(lr){3-8}
& & \textit{ER ($\tau$ = 64)} & 45.66 & 46.36 & 48.50 & 51.04 & 53.51 \\
& & & \textcolor{gray}{\footnotesize 0.53} & \textcolor{gray}{\footnotesize 1.37} & \textcolor{gray}{\footnotesize 7.91} & \textcolor{gray}{\footnotesize 24.81} & \textcolor{gray}{\footnotesize 100.61} \\
\cmidrule(lr){3-8}
& & \textit{ER ($\tau$ = 128)} & 47.13 & 48.54 & 50.91 & 53.11 & 56.84 \\
& & & \textcolor{gray}{\footnotesize 0.49} & \textcolor{gray}{\footnotesize 1.12} & \textcolor{gray}{\footnotesize 5.76} & \textcolor{gray}{\footnotesize 17.33} & \textcolor{gray}{\footnotesize 79.98} \\

\cmidrule(lr){2-8}

% Qwen2.5-3b with Skywork-1.5b
& \multirow{8}{*}{\begin{tabular}[c]{@{}c@{}}\textbf{Skywork}\\\textbf{-1.5b}\end{tabular}} 
& \textit{Vanilla} & 31.63 & 40.49 & 44.51 & 47.29 & 50.98 \\
& & & \textcolor{gray}{\footnotesize 1.37} & \textcolor{gray}{\footnotesize 4.77} & \textcolor{gray}{\footnotesize 10.37} & \textcolor{gray}{\footnotesize 27.31} & \textcolor{gray}{\footnotesize 88.77} \\
\cmidrule(lr){3-8}
& & \textit{ER ($\tau$ = 32)} & 37.13 & 43.13 & 45.19 & 49.59 & 51.33 \\
& & & \textcolor{gray}{\footnotesize 0.33} & \textcolor{gray}{\footnotesize 1.36} & \textcolor{gray}{\footnotesize 6.67} & \textcolor{gray}{\footnotesize 17.29} & \textcolor{gray}{\footnotesize 47.43} \\
\cmidrule(lr){3-8}
& & \textit{ER ($\tau$ = 64)} & 40.67 & 43.26 & 47.88 & 51.41 & 53.88 \\
& & & \textcolor{gray}{\footnotesize 0.31} & \textcolor{gray}{\footnotesize 1.28} & \textcolor{gray}{\footnotesize 6.40} & \textcolor{gray}{\footnotesize 15.95} & \textcolor{gray}{\footnotesize 42.45} \\
\cmidrule(lr){3-8}
& & \textit{ER ($\tau$ = 128)} & 42.26 & 46.55 & 51.82 & 52.61 & 55.09 \\
& & & \textcolor{gray}{\footnotesize 0.25} & \textcolor{gray}{\footnotesize 0.60} & \textcolor{gray}{\footnotesize 2.40} & \textcolor{gray}{\footnotesize 7.50} & \textcolor{gray}{\footnotesize 25.33} \\

\bottomrule
\end{tabular}
\end{adjustbox}
\end{table}

Table~\ref{tab:Early_rejection_performance} extends the analysis to the Math-500 and AIME 2024 benchmarks, using MathShepherd-Mistral-7B as the PRM. Again, we observe consistent trends across datasets: as $\tau$ increases, early rejection becomes more selective and cost-efficient, with only minor losses (if any) in final accuracy.

\begin{table}[htbp]
\centering
\caption{Results on Math-500 and AIME datasets with MathShepherd-Mistral-7B as the PRM. Each configuration shows accuracy (top) and total FLOPs (bottom) for different beam sizes and $\tau$ thresholds.}
\label{tab:Early_rejection_performance}
\begin{adjustbox}{max width=\textwidth}
\begin{tabular}{@{}l|l|l|*{5}{c}@{}}
\toprule
\multirow{2}{*}{\textbf{Dataset}} & \multirow{2}{*}{\textbf{Model}} & \multirow{2}{*}{\textbf{Setting}} & \multicolumn{5}{c}{\textbf{Number of Samples ($\tau$)}} \\
\cmidrule(lr){4-8}
& & & \textbf{4} & \textbf{8} & \textbf{16} & \textbf{32} & \textbf{64} \\
\midrule

% Math-500 - Llama-3.2-3b
\multirow{8}{*}{\rotatebox{90}{\textbf{Math-500}}} 
& \multirow{8}{*}{\begin{tabular}[c]{@{}c@{}}\textbf{Llama-3.2}\\\textbf{-3b}\end{tabular}} 
& \textit{Vanilla} & 46.20 & 48.00 & 49.06 & 50.81 & 51.44 \\
& & & \textcolor{gray}{\footnotesize 5.04} & \textcolor{gray}{\footnotesize 27.51} & \textcolor{gray}{\footnotesize 33.22} & \textcolor{gray}{\footnotesize 137.54} & \textcolor{gray}{\footnotesize 202.27} \\
\cmidrule(lr){3-8}
& & \textit{ER ($\tau$ = 32)} & 39.63 & 40.30 & 44.60 & 46.60 & 47.21 \\
& & & \textcolor{gray}{\footnotesize 1.68} & \textcolor{gray}{\footnotesize 10.15} & \textcolor{gray}{\footnotesize 27.42} & \textcolor{gray}{\footnotesize 92.15} & \textcolor{gray}{\footnotesize 189.23} \\
\cmidrule(lr){3-8}
& & \textit{ER ($\tau$ = 64)} & 42.00 & 43.20 & 48.67 & 50.43 & 51.19 \\
& & & \textcolor{gray}{\footnotesize 1.50} & \textcolor{gray}{\footnotesize 8.67} & \textcolor{gray}{\footnotesize 23.45} & \textcolor{gray}{\footnotesize 101.17} & \textcolor{gray}{\footnotesize 184.71} \\
\cmidrule(lr){3-8}
& & \textit{ER ($\tau$ = 128)} & 45.46 & 46.80 & 48.74 & 50.29 & 51.34 \\
& & & \textcolor{gray}{\footnotesize 0.60} & \textcolor{gray}{\footnotesize 3.21} & \textcolor{gray}{\footnotesize 18.91} & \textcolor{gray}{\footnotesize 77.46} & \textcolor{gray}{\footnotesize 138.63} \\

\cmidrule(lr){2-8}

% Math-500 - Qwen2.5-3b
& \multirow{8}{*}{\begin{tabular}[c]{@{}c@{}}\textbf{Qwen2.5}\\\textbf{-3b}\end{tabular}} 
& \textit{Vanilla} & 51.67 & 53.25 & 54.08 & 56.73 & 58.80 \\
& & & \textcolor{gray}{\footnotesize 14.02} & \textcolor{gray}{\footnotesize 47.48} & \textcolor{gray}{\footnotesize 65.32} & \textcolor{gray}{\footnotesize 250.03} & \textcolor{gray}{\footnotesize 536.10} \\
\cmidrule(lr){3-8}
& & \textit{ER ($\tau$ = 32)} & 45.87 & 49.59 & 51.41 & 52.80 & 55.60 \\
& & & \textcolor{gray}{\footnotesize 2.41} & \textcolor{gray}{\footnotesize 10.58} & \textcolor{gray}{\footnotesize 56.49} & \textcolor{gray}{\footnotesize 134.12} & \textcolor{gray}{\footnotesize 354.91} \\
\cmidrule(lr){3-8}
& & \textit{ER ($\tau$ = 64)} & 53.88 & 54.19 & 55.60 & 57.11 & 59.34 \\
& & & \textcolor{gray}{\footnotesize 2.10} & \textcolor{gray}{\footnotesize 9.28} & \textcolor{gray}{\footnotesize 42.33} & \textcolor{gray}{\footnotesize 112.46} & \textcolor{gray}{\footnotesize 263.08} \\
\cmidrule(lr){3-8}
& & \textit{ER ($\tau$ = 128)} & 55.21 & 59.43 & 60.40 & 62.61 & 66.13 \\
& & & \textcolor{gray}{\footnotesize 1.61} & \textcolor{gray}{\footnotesize 7.45} & \textcolor{gray}{\footnotesize 32.54} & \textcolor{gray}{\footnotesize 94.52} & \textcolor{gray}{\footnotesize 195.23} \\

\midrule

% AIME - Llama-3.2-3b
\multirow{8}{*}{\rotatebox{90}{\textbf{AIME}}} 
& \multirow{8}{*}{\begin{tabular}[c]{@{}c@{}}\textbf{Llama-3.2}\\\textbf{-3b}\end{tabular}} 
& \textit{Vanilla} & 3.33 & 6.67 & 6.67 & 10.00 & 13.33 \\
& & & \textcolor{gray}{\footnotesize 0.10} & \textcolor{gray}{\footnotesize 0.25} & \textcolor{gray}{\footnotesize 0.72} & \textcolor{gray}{\footnotesize 1.56} & \textcolor{gray}{\footnotesize 2.61} \\
\cmidrule(lr){3-8}
& & \textit{ER ($\tau$ = 32)} & 0.00 & 3.33 & 3.33 & 6.67 & 10.00 \\
& & & \textcolor{gray}{\footnotesize 0.05} & \textcolor{gray}{\footnotesize 0.16} & \textcolor{gray}{\footnotesize 0.46} & \textcolor{gray}{\footnotesize 1.14} & \textcolor{gray}{\footnotesize 2.13} \\
\cmidrule(lr){3-8}
& & \textit{ER ($\tau$ = 64)} & 3.33 & 3.33 & 10.00 & 10.00 & 13.33 \\
& & & \textcolor{gray}{\footnotesize 0.02} & \textcolor{gray}{\footnotesize 0.09} & \textcolor{gray}{\footnotesize 0.41} & \textcolor{gray}{\footnotesize 0.72} & \textcolor{gray}{\footnotesize 1.89} \\
\cmidrule(lr){3-8}
& & \textit{ER ($\tau$ = 128)} & 3.33 & 6.67 & 10.00 & 13.33 & 13.33 \\
& & & \textcolor{gray}{\footnotesize 0.02} & \textcolor{gray}{\footnotesize 0.04} & \textcolor{gray}{\footnotesize 0.38} & \textcolor{gray}{\footnotesize 0.61} & \textcolor{gray}{\footnotesize 2.01} \\

\cmidrule(lr){2-8}

% AIME - Qwen2.5-3b
& \multirow{8}{*}{\begin{tabular}[c]{@{}c@{}}\textbf{Qwen2.5}\\\textbf{-3b}\end{tabular}} 
& \textit{Vanilla} & 6.67 & 10.00 & 10.00 & 13.33 & 16.67 \\
& & & \textcolor{gray}{\footnotesize 0.13} & \textcolor{gray}{\footnotesize 0.31} & \textcolor{gray}{\footnotesize 1.19} & \textcolor{gray}{\footnotesize 2.68} & \textcolor{gray}{\footnotesize 5.51} \\
\cmidrule(lr){3-8}
& & \textit{ER ($\tau$ = 32)} & 3.33 & 6.67 & 6.67 & 10.00 & 10.00 \\
& & & \textcolor{gray}{\footnotesize 0.05} & \textcolor{gray}{\footnotesize 0.21} & \textcolor{gray}{\footnotesize 0.63} & \textcolor{gray}{\footnotesize 1.34} & \textcolor{gray}{\footnotesize 3.35} \\
\cmidrule(lr){3-8}
& & \textit{ER ($\tau$ = 64)} & 6.67 & 6.67 & 10.00 & 13.33 & 13.33 \\
& & & \textcolor{gray}{\footnotesize 0.04} & \textcolor{gray}{\footnotesize 0.12} & \textcolor{gray}{\footnotesize 0.47} & \textcolor{gray}{\footnotesize 0.93} & \textcolor{gray}{\footnotesize 2.36} \\
\cmidrule(lr){3-8}
& & \textit{ER ($\tau$ = 128)} & 6.67 & 6.67 & 10.00 & 13.33 & 16.67 \\
& & & \textcolor{gray}{\footnotesize 0.02} & \textcolor{gray}{\footnotesize 0.09} & \textcolor{gray}{\footnotesize 0.39} & \textcolor{gray}{\footnotesize 0.77} & \textcolor{gray}{\footnotesize 2.12} \\

\bottomrule
\end{tabular}
\end{adjustbox}
\end{table}

Table~\ref{tab:flops_comparison} aggregates FLOP consumption across all LLM–PRM combinations under three decoding regimes: Vanilla, ER($\tau{=}32$), and ER($\tau{=}64$). The results reveal that early rejection with $\tau = 64$ consistently achieves the lowest compute cost without compromising output quality, yielding up to 9× reduction in total inference FLOPs.

Together, these tables validate the scalability and robustness of our early rejection method across models, evaluators, datasets, and rejection thresholds.

\begin{table}[htbp]
\centering
\caption{Total FLOPs ($\times 10^{18}$) for each LLM–PRM combination under vanilla decoding and early rejection at $\tau=32$ and $\tau=64$. Early rejection consistently reduces compute, with Qwen-based configurations showing the largest savings.}
\label{tab:flops_comparison}
\begin{tabular}{@{}lcccccc@{}}
\toprule
\multirow{2}{*}{\textbf{Model Combination}} & \multicolumn{2}{c}{\textbf{Vanilla}} & \multicolumn{2}{c}{\textbf{Early Rejection ($\tau$=32)}} & \multicolumn{2}{c}{\textbf{Early Rejection ($\tau$=64)}} \\
\cmidrule(lr){2-3} \cmidrule(lr){4-5} \cmidrule(lr){6-7}
& \textbf{LLM} & \textbf{PRM} & \textbf{LLM} & \textbf{PRM} & \textbf{LLM} & \textbf{PRM} \\
\midrule
Llama+Math    & 3.70  & 27.51 & 5.73  & 16.27 & 4.62  & 10.23 \\
Llama+Skywork & 19.79 & 6.06  & 7.54  & 2.29  & 5.67  & 1.96  \\
Qwen+Math     & 13.22 & 67.19 & 7.46  & 18.27 & 7.54  & 17.27 \\
Qwen+Skywork  & 19.22 & 8.08  & 12.17 & 5.12  & 10.89 & 5.06  \\
\bottomrule
\end{tabular}
\end{table}

\end{document}